%% file: main.tex
\documentclass[runningheads]{llncs}
\usepackage{eccv}
\usepackage{eccvabbrv}
\usepackage{graphicx}
\usepackage{booktabs}
\usepackage{multirow}
\usepackage{colortbl}
\usepackage{appendix}
\definecolor{Gray}{gray}{0.9}
\definecolor{Q-Gray}{gray}{0.95}
\usepackage[accsupp]{axessibility}  
\usepackage{algorithm} 
\usepackage{algorithmic}
\usepackage{hyperref}

\usepackage{orcidlink}
\def\Ie{\emph{I.e}\onedot}
\def\cf{\emph{c.f}\onedot} \def\Cf{\emph{C.f}\onedot}
\def\etc{\emph{etc}\onedot} \def\vs{\emph{vs}\onedot}
\def\wrt{w.r.t\onedot} 
\def\aka{a.k.a\onedot} 
\def\dof{d.o.f\onedot}
\def\etal{\emph{et al}\onedot}
\def\ie{\textit{i.e.}}
\def\eg{\textit{e.g.}}

\begin{document}

\title{Bridge Past and Future: Overcoming Information Asymmetry in Incremental Object Detection} 
\titlerunning{Bridge Past and Future}
\author{Qijie Mo\inst{1,3}\orcidlink{0009-0005-6704-685X} \and%
Yipeng Gao\inst{1,3} \orcidlink{0009-0009-6784-6944} \and
Shenghao Fu\inst{1,3} \orcidlink{0009-0008-3456-4193} \and
Junkai Yan\inst{1,3}\orcidlink{0009-0009-6531-0070} \and \\
Ancong Wu\inst{1,3,}\thanks{Corresponding Author}\orcidlink{0000-0002-7969-3190}\and
Wei-Shi Zheng\inst{1,2,3,}$^\star$\orcidlink{0000-0001-8327-0003} 
}

\authorrunning{Q.~Mo et al.}
\institute{
School of Computer Science and Engineering, Sun Yat-sen University, China \and
Peng Cheng Laboratory, Shenzhen, China \and
Key Laboratory of Machine Intelligence and Advanced Computing, \\ Ministry of Education, China \\
\email{\{moqj3,gaoyp23,fushh7,yanjk3\}@mail2.sysu.edu.cn},\\ 
\email{wuanc@mail.sysu.edu.cn}, \email{wszheng@ieee.org}\\
}

\maketitle

\begin{abstract}
    In incremental object detection, knowledge distillation has been proven to be an effective way to alleviate catastrophic forgetting.
    However, previous works focused on preserving the knowledge of old models, ignoring that images could simultaneously contain categories from past, present, and future stages. The co-occurrence of objects makes the optimization objectives inconsistent across different stages since the definition for foreground objects differs across various stages, which limits the model's performance greatly.
    To overcome this problem, we propose a method called ``Bridge Past and Future'' (BPF), which aligns models across stages, ensuring consistent optimization directions.
    In addition, we propose a novel Distillation with Future (DwF) loss, fully leveraging the background probability to mitigate the forgetting of old classes while ensuring a high level of adaptability in learning new classes.
    Extensive experiments are conducted on both Pascal VOC and MS COCO benchmarks. Without memory, BPF outperforms current state-of-the-art methods under various settings. The code is available at \url{https://github.com/iSEE-Laboratory/BPF}.
\end{abstract}

\section{Introduction}
\label{sec:intro}

Object detection~\cite{ren2015faster,detr, fu2023asag, deformabledetr,lin2017focal,tian2019fcos} is a fundamental computer vision task and has significantly given rise to many sub-directions~\cite{acrofod, asyfod, few-shot_OD,few-shot_OD2,OVOD1,OVOD2,cao2023contrastive,10219835}. 
General detection models typically assume full access to all interest classes during training and require exhaustive labeled data from the start. 
However, in dynamic real-world applications, where new object categories may appear continually, object detection models are expected to adapt to these ongoing changing classes, especially when there is limited storage to save all the data from the beginning to the present or when data privacy concerns are encountered. 
To this end, incremental object detection (IOD)~\cite{joseph2021incremental,liu2023augmented,zhou2020lifelong,yang2022multi,zohar2023prob,cermelli2022modeling} has caught progressive attention, which aims to continually detect new objects without forgetting the previously learned ones.
\begin{figure}[tb]
  \centering
  \includegraphics[width=1.0\linewidth]{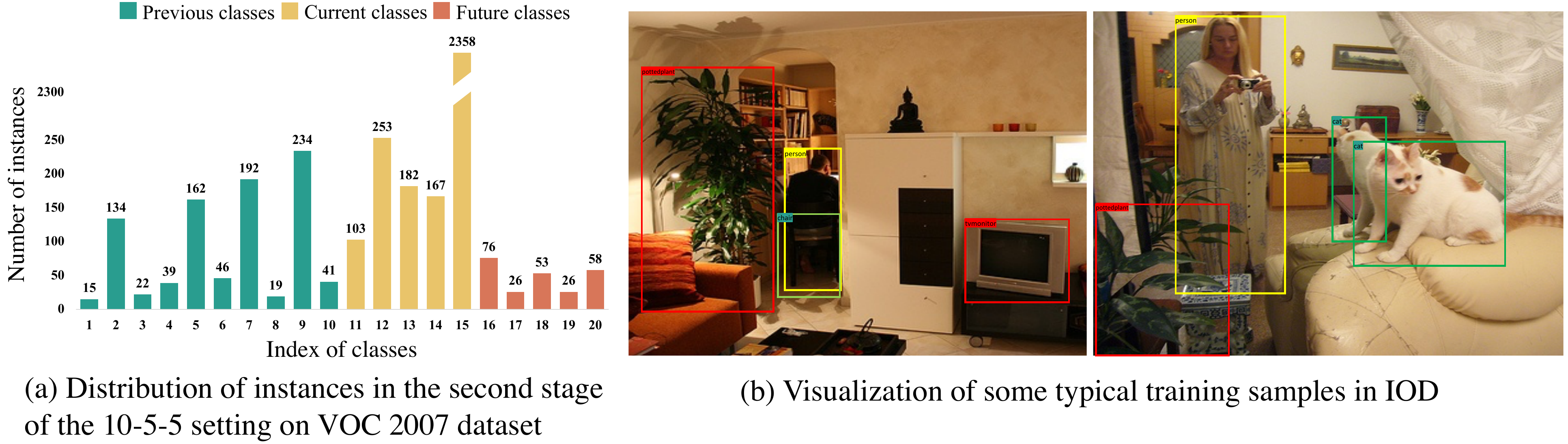}
  \caption{Challenges faced by the IOD task. Classes from \textcolor{green}{previous}, \textcolor{yellow}{current}, and potential \textcolor{red}{future} stages may appear arbitrarily in the current training stage, while only the annotations of current classes are available to train the detector. Best viewed in color.}
  \label{fig:explain}
\end{figure}

Unlike incremental learning in classification, the incremental object detection task faces a dual challenge caused by the concurrence of different classes: (1) Foreground objects identified in previous stages are \textbf{unlabeled} in the current stage, and thus, the model currently trained may regard them as background. (2) The current stage's background may also contain objects that will be recognized as the foreground in future stages.
Taking the 10-5-5 setting on the VOC dataset\cite{everingham2010pascal} as an example, as shown in \Cref{fig:explain}, a large proportion of instances belonging to classes of the previous and future stages occur concurrently in the training images in the current stage. 
This asymmetry in information between the past and current stages causes the current model to erroneously classify objects of old categories as background, aggravating the catastrophic forgetting problem.
Similarly, the asymmetry between the current and future stages leads the current model to classify objects of potential future categories as background erroneously, which requires models trained in future stages to rectify their misperception, thus hindering the learning of new classes. 
As a result, the optimization objectives across different learning stages are inconsistent, significantly limiting the generalization of current incremental detectors.
Previous works~\cite{zhou2020lifelong, yang2022multi, cermelli2022modeling} concentrate on introducing strong regularization to prevent catastrophic forgetting but ignore the impact of concurrence of different classes, resulting in limited performance on both old and new classes.

This work aims to tackle the critical information asymmetry challenge in IOD by utilizing the abundant concurrence information within an image. 
We introduce a novel approach named Bridge Past and Future (BPF), which is designed to connect the past and future stages with the current stage, ensuring the model adheres to consistent optimization objectives throughout the entire incremental learning process. 
For the past stages, we utilize the previous model as a reliable labeler to construct some pseudo labels for old classes and combine them with current annotations for training the current model symmetrically.
Simultaneously, we re-identify some salient regions that may contain potential future objects out of the background and exclude them from negative samples in the current stage to avoid classifying them into the background at this stage and disturbing future learning, which is expected to achieve symmetry with future stages. 
These two novel designs alleviate the impact of inconsistent training objectives across different stages from the aspects of the past and the future, thus easing the training difficulty of incremental detectors.

In addition, we propose a distillation technique to augment the detector in the current stage, where we take a step back by viewing the current stage as a clear future stage for previous stages and propose a Distillation with Future~(DwF) loss. 
The distillation is carried out by two teachers, \ie, the detector trained on the previous stage and an expert detector trained with only current data. 
The student detector absorbs the knowledge of old and current classes from the old and expert models, respectively, in a class-by-class manner, preventing the detector from catastrophic forgetting and facilitating learning of current classes.

We conduct extensive experiments on the PASCAL VOC~\cite{everingham2010pascal} and COCO~\cite{lin2014microsoft} datasets to evaluate the effectiveness of our BPF, which outperforms other state-of-the-art (SOTA) methods under multiple settings in a memory-free way. Moreover, we also conduct comprehensive ablation studies and visualizations to help better understand how each component of the BPF works.

\section{Related Works}
\label{sec:related-work}
 
\textbf{Object Detection.}
Traditional object detectors can be primarily classified into two streams: one-stage  \cite{tian2019fcos,tan2020efficientdet,redmon2018yolov3,lin2017focal} and two-stage detectors\cite{he2017mask,lin2017feature,ren2015faster,girshick2015fast}. Two-stage detectors first predict several coarse candidate proposals via region proposal extractors~\cite{ren2015faster,girshick2015fast} and then adopt a region of interest (RoI) head to refine these proposals and output final predictions. Unlike two-stage detectors, one-stage counterparts directly generate final outputs without predicting candidate proposals. Despite performing well under the standard training setup, both of them fail to generalize to the incremental training setup due to lacking previous training data. Without losing generality, this paper concentrates on enhancing two-stage detectors, \ie, Faster R-CNN\cite{ren2015faster}, enabling it to learn new classes incrementally while retaining previously acquired knowledge.

\vspace{0.5em}
\noindent \textbf{Incremental Learning.}
Incremental learning methods can be mainly divided into three categories: memory-based\cite{tang2022learning,castro2018end,hou2019learning,ostapenko2019learning,rebuffi2017icarl,shin2017continual,wu2018memory,li2022class,li2024continual}, 
regularization-based\cite{chaudhry2018riemannian,dhar2019learning,kirkpatrick2017overcoming,li2017learning,zenke2017continual}, 
and structure-based\cite{tang2023when,mallya2018piggyback,mallya2018packnet,rusu2016progressive,liu2021l3doc}.
Memory-based methods store a handful of samples for replaying~\cite{liu2023augmented,liu2023continual,hou2019learning} or suggest generating samples\cite{kemker2017fearnet,wu2018memory} for data compensation in new stages.
Some regularization-based methods try to eliminate the effect of new tasks on previously learned knowledge by identifying the most crucial parameters\cite{lopez2017gradient,wu2018memory,kirkpatrick2017overcoming}, while some other regularization-based methods propose to distill knowledge\cite{hinton2015distilling} from old to current models by transiting knowledge on logits, final features or intermediate features\cite{li2017learning,hou2019learning,douillard2022dytox,douillard2020podnet,dhar2019learning}.
Structure-based methods dedicate specific parameters to each task, freezing them to mitigate forgetting.
In this study, we focus on the regularization-based knowledge distillation approach for object detection and introduce a novel distillation with future information.

\vspace{0.5em}
\noindent \textbf{Incremental Object Detection.}
Most previous works\cite{kang2023alleviating,liu2023continual,peng2023diode,shieh2020continual,li2019rilod,dong2023class} focus on distilling old classes' knowledge of intermediate features\cite{cermelli2022modeling,liu2023augmented,peng2020faster,yang2022multi,zhou2020lifelong,liu2020multi,hao2019end}, region proposal network\cite{cermelli2022modeling,peng2020faster,zhou2020lifelong}, or RoI head\cite{feng2022overcoming} to prevent forgetting. 
A stream of methods also pays attention to classification loss. PPAS\cite{zhou2020lifelong} introduces a pseudo-positive-aware sampling algorithm to identify regions corresponding to old classes and prevents them from being sampled as background. MMA\cite{cermelli2022modeling} proposes unbiased classification loss, consolidating the background and all old classes into one entity, which aims to minimize the optimization objective conflict between the current stage's background and the past stages' foreground. However, it diminishes the ability to distinguish old classes from the background.
Other methods\cite{acharya2020rodeo, gupta2022ow, joseph2021incremental, joseph2021towards, liu2023augmented, yang2023pseudo} focus on rehearsal to maintain the previous stage knowledge, either performing replay of the intermediate features\cite{acharya2020rodeo}, the images\cite{liu2023continual,joseph2021incremental,joseph2021towards}, or the instances\cite{liu2023augmented}.
Incdet\cite{liu2020incdet} presents a parameter isolation strategy that builds upon EWC\cite{kirkpatrick2017overcoming}.
In this study, in addition to preserving knowledge for old classes, we also take future classes into account, aligning optimization objectives consistently across all stages.


\section{Method}

\subsection{Preliminaries}
\noindent\textbf{Problem Formulation.}
\label{subsec: problem-formulation}
In incremental object detection, the training is performed over multiple learning stages, each one introducing a new set of classes to be detected.
Let $\mathcal{C}$ denote the set of classes that are incrementally introduced to the object detector $\mathcal{M}$.
In the $t$-th training stage, a grouping of classes $\mathcal{C}_t$ are introduced to the detector: $\mathcal{C}_t \subset \mathcal{C}$, such that $\mathcal{C}_i \cap \mathcal{C}_j = \varnothing$, for any $i \ne j$ and $i,j\leq t$.
Let $\mathcal{D}_t$ denote the images containing annotated objects of classes in $\mathcal{C}_t$.
Each image can contain multiple objects of different classes, but annotations $\mathcal{Y}_t$ are available only for those object instances that belong to classes in $\mathcal{C}_t$.
The challenge in class incremental object detection is to continually update $\mathcal{M}_{t}$ to $\mathcal{M}_{t+1}$ by learning $\mathcal{C}_{t+1}$, without access to $\{\mathcal{D}_0,\cdots,\mathcal{D}_t \}$ while maintaining original performance on $\{\mathcal{C}_0,\cdots,\mathcal{C}_t\}$.

\vspace{0.5em}
\noindent\textbf{General Detector Training Pipeline.}
This work starts with the representative Faster RCNN-like detectors~\cite{ren2015faster} $f=\{f_b,f_{rpn},f_{roi}\}$, which generally includes a backbone network $f_b$, Region Proposal Network (RPN) $f_{rpn}$ and Region of Interest Head (RoI Head) $f_{roi}$.
During training, all training proposals are categorized into positive, negative, and ignored samples for supervised training based on their IoUs with the ground truth. Positive samples are assigned to predict the class of their matched ground truth, whereas negative samples are designated as background. Thus, the objects without annotations in the current stage will be classified as background, hindering the learning of incremental models.

\vspace{0.5em}
\noindent\textbf{Common Distillation Methods for Incremental Object Detection.}
To prevent forgetting in IOD, a widely adopted solution involves knowledge distillation in RPN and RoI head\cite{shmelkov2017incremental,peng2020faster,zhou2020lifelong,cermelli2022modeling}:
\begin{equation}
    \mathcal{L}_{dist} = \mathcal{L}_{dist}^{rpn} + \mathcal{L}_{dist}^{roi}.
\end{equation}
The distillation in the RoI head includes the L2 loss between the box coordinates and the Kullback-Leibler divergence for class probabilities. MMA\cite{cermelli2022modeling} notices the missing old annotations and proposes Unbiased Knowledge Distillation (UKD):
\begin{equation}
    \mathcal{L}_{dist,cls}^{roi}(i)= \frac{1}{|\mathcal{C}_{t-1}|+1}(p_i^{b,t-1}\log(p_i^{b,t}+\sum_{c\in\mathcal{C}_t}p_i^{c,t})+\sum_{c\in\mathcal{C}_{1:t-1}}p_i^{c,t-1}\log(p_i^{c,t})),
\end{equation}
where $p_i^{c,t-1}$ and $p_i^{c,t}$ indicate the classification output for the proposal $i$ and class $c$ of the teacher (old) model and student (current) model, respectively, and $b$ is the background class. As a common practice, 64 proposals are used for distillation, which are randomly selected out of the top 128 proposals with the highest objectness scores from the RPN network of the old model. However, given the class-agnostic characteristic of the RPN, the proposals for distillation cover objects of both old and current classes. In such cases, the old model, without seeing current classes, cannot provide useful knowledge for the current classes and even hinders their learning. Further, treating the background probability and current class probabilities as a unified entity during distillation inevitably diminishes the ability to differentiate between current classes and background.

\subsection{Overall Framework}
Unlike classification tasks with a single label per input, incremental object detection presents a scenario where an image $\mathcal{I}_t$ can encompass objects from the current class set $\mathcal{C}_t$, previous class sets $\mathcal{C}_{1:t-1}$ and future class sets $\mathcal{C}_{t+1:\infty}$.
Nonetheless, the annotations $\mathcal{Y}_t$ are limited to bounding boxes and class labels for objects in $\mathcal{C}_t$, while objects of other classes are regarded as background.

\begin{figure}[t]
    \centering
    \includegraphics[width=0.9\linewidth]{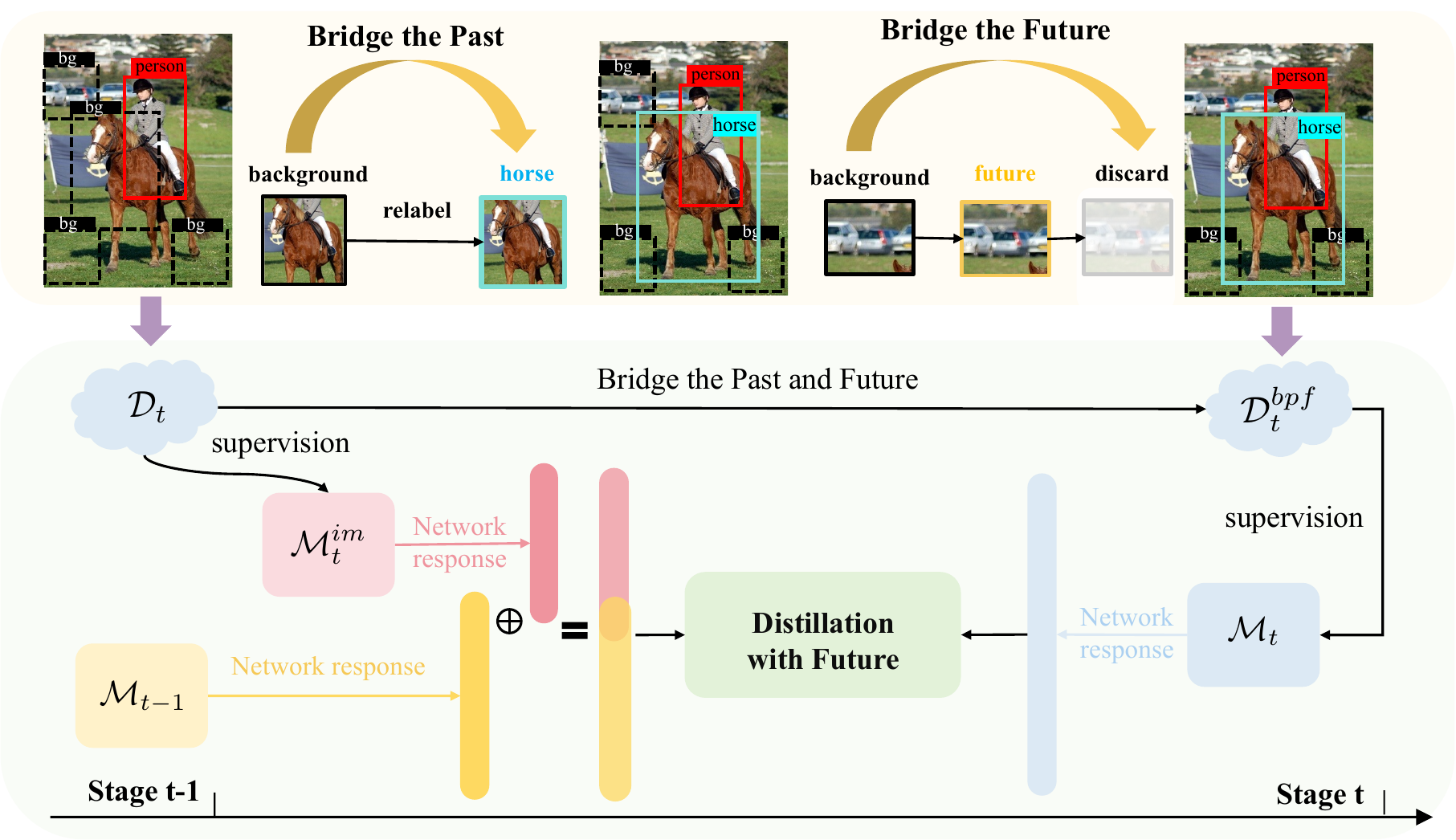}
    \caption{The overall framework of our method. The top side illustrates the Bridge Past and Future (BPF) procedure, which identifies objects of past classes and excludes several potential objects of future classes to ensure consistent optimization during the entire training process. The bottom side shows the Distillation with Future (DwF) process, which employs both the old model $\mathcal{M}_{t-1}$ adept at detecting old categories and the interim model $\mathcal{M}_{t}^{im}$ trained on $\mathcal{D}_t$ and specialized in new categories, to conduct a comprehensive distillation across all categories for the current model $\mathcal{M}_t$.}
    \label{fig:overall}
\end{figure}

To address the challenge of inconsistent optimization objectives, we introduce a novel strategy termed Bridge Past and Future (BPF). As shown in \Cref{fig:overall}, the method is divided into two parts. From the perspective of supervised learning, we use high-confidence predictions of old classes given by the old model as pseudo labels to bridge the information from the past stage (\Cref{sec:bridge_past}) and discard some potential objects from the background to bridge the future stage (\Cref{sec:bridge_future}). Regarding the distillation learning, we propose a novel Distillation with Future (DwF) loss by considering current classes in the distillation. The distillation probabilities are combined from two teacher models: the old model $\mathcal{M}_{t-1}$ for old classes and an expert model for current classes. By utilizing abundant information in background probability, the distillation takes both previous stages and the current stage into consideration (\Cref{sec:dwf_loss}).

\begin{figure}[t]
    \centering
    \includegraphics[width=0.8\linewidth]{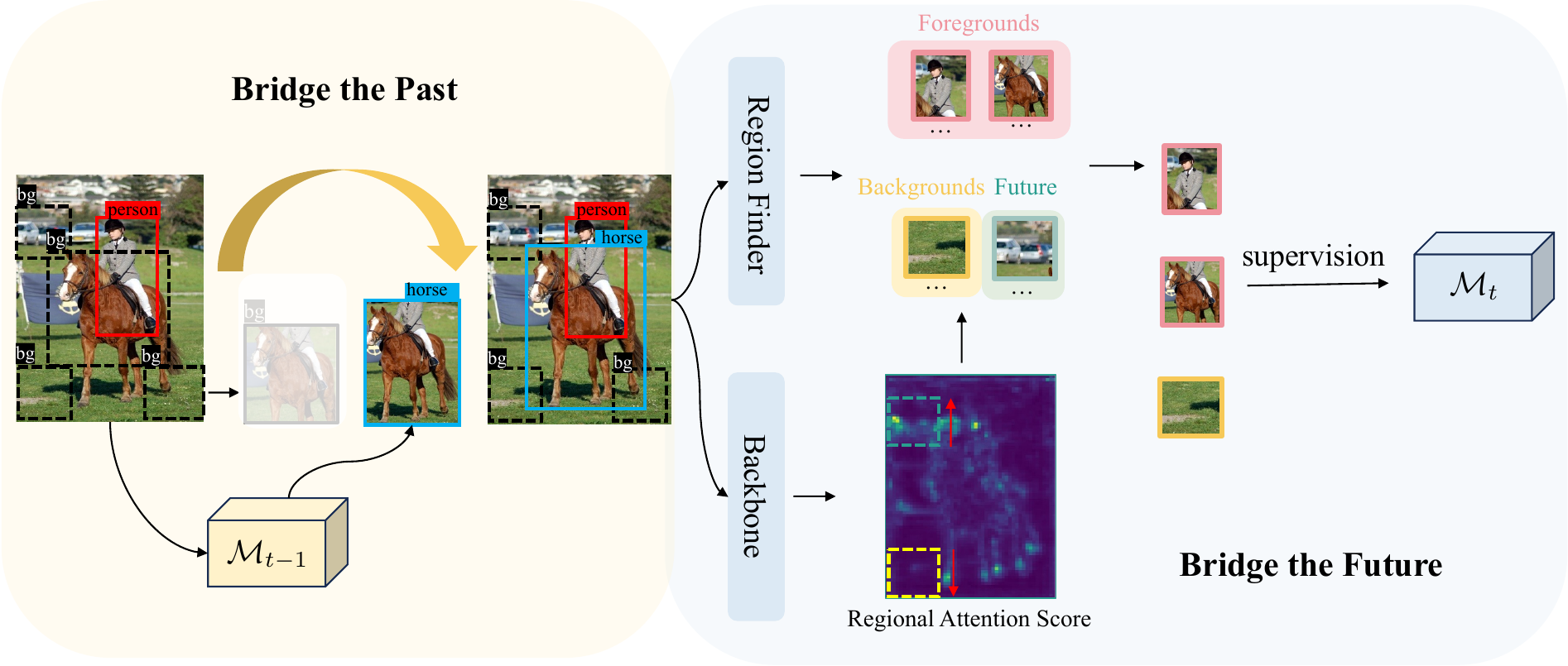}
    \caption{Overview of Bridge Past and Future. We adopt the previous model $\mathcal{M}_{t-1}$ to predict some pseudo labels for past classes to complement their missing supervision in the current stage. Additionally, we exclude several proposals that are likely to be an object but are not included in the current ground truth and pseudo labels from the background to avoid classifying them into background mistakenly.}
    \label{fig:BPF}
\end{figure}

\subsection{Bridging the Past}
\label{sec:bridge_past}

To address the inconsistency in optimization objectives between the previous and current models, we bridge the past information to the current stage by incorporating pseudo supervision signals from past stages into the current model's optimization process.
The old model was trained under the supervision of human annotations, which is a high-quality pseudo labeler for old classes. 
We take the high-confidence inference results of the old model $\mathcal{M}_{t-1}$ on current training images as pseudo labels and combine them with current annotations to train the current detector, thereby aligning the supervision of old classes with previous stages.
The left part of \Cref{fig:BPF} shows the pipeline of bridging the past strategy.

Formally, given predictions $\hat{y}^{old} \in \mathcal{P}$ of the old model $\mathcal{M}_{t-1}$ on current images, we first select some high-confidence regions of previous classes $\mathcal{C}_{1:t-1}$ with a confidence threshold $\eta$, followed by the Non-Maximum Suppression (NMS) operation to reduce duplication:
\begin{equation}
    \mathcal{U} = \{j \in \mathcal{P}: \max_{c \in \mathcal{C}_{1:t-1}} \hat{p}_j^{\mathrm{old}}(c) > \eta \},\quad 
    \mathcal{E} = \mathrm{NMS}(\mathcal{U}).
\end{equation}
Then, we further narrow down the predictions to a subset $\mathcal{W} \subset \mathcal{E}$ that do not overlap with the ground-truth labels of the new categories via an IoU threshold $\lambda_1$, ensuring a clear distinction between past and present classes:
\begin{equation}
    \mathcal{W}=\{j\in\mathcal{E}:\forall i\in\mathcal{Y}_t,\mathrm{IoU}(\hat{\boldsymbol{b}}_j^{\mathrm{old}},\boldsymbol{b}_i)\leq \lambda_1\}.
    \label{eq:pseudo-labels}
\end{equation}

By modeling the previous supervision signals from the old model, we bridge the past stages to the current one, ensuring the current model's optimization direction encompasses the objectives of earlier stages, significantly mitigating the forgetting problem. 
MMA\cite{cermelli2022modeling} treats the background probability and old class probabilities as a unified entity in the classification loss, making old classes hard to separate from the background. While our method explicitly models the old classes in the current stage out of the background.

\subsection{Bridging the Future}
\label{sec:bridge_future}

Bridging the future aims to rectify the misalignment between the optimization goals of the current and upcoming stages, as objects from future class sets $\mathcal{C}_{t+1:\infty}$ in the current dataset $\mathcal{D}_t$ are classified as background. The core design is to find some salient objects in the background and exclude them from the negative samples during the training. By separating pure background from regions likely to contain future category objects in the current stage, we align the optimization objectives for background treatment across both current and future models.

Specifically, we find that the activation in feature maps is a good indicator for salient objects. As illustrated in \Cref{fig:BPF}, in the absence of annotations for old and future classes, foreground and background features demonstrate notable differences in spatial attention.
We produce the attention map $A_i \in \mathbb{R}^{H\times W}$ from the backbone feature map $F_i = f_b(I_i) \in \mathbb{R}^{H\times W \times C}$: 
\begin{equation}
    A_i = \mathrm{Softmax}\left(\sum_{c=1}^C |F_i|^p \right),
\end{equation}
where $H$, $W$, and $C$ denote the feature’s height, width, and channel.
We compute the attention score for each region, indicating the likelihood of containing foreground objects. The attention score $a_{i,j}^{roi}$ for region $r_j$ is calculated as follows:
\begin{equation}
    a_{i,j}^{roi} = \mathrm{Avg}(\mathrm{RoIPool}(A_i,r_j)) \in \mathbb{R}.
\end{equation}
Regions with high attention scores $a_{i,j}^{roi}$ from feature maps and objectness scores $o_j$ from class-agnostic RPN imply a greater chance of being future category objects. We discard these RoIs when sampling negative samples when training the RoI head, thus maintaining the model's consistency with future stage background definitions. Regions with lower scores or having a considerable IoU with ground truth are considered reliable backgrounds.

\subsection{Combining Current Classes into Distillation}
\label{sec:dwf_loss}

\begin{figure}[t]
    \centering
    \includegraphics[width=1.0\linewidth]{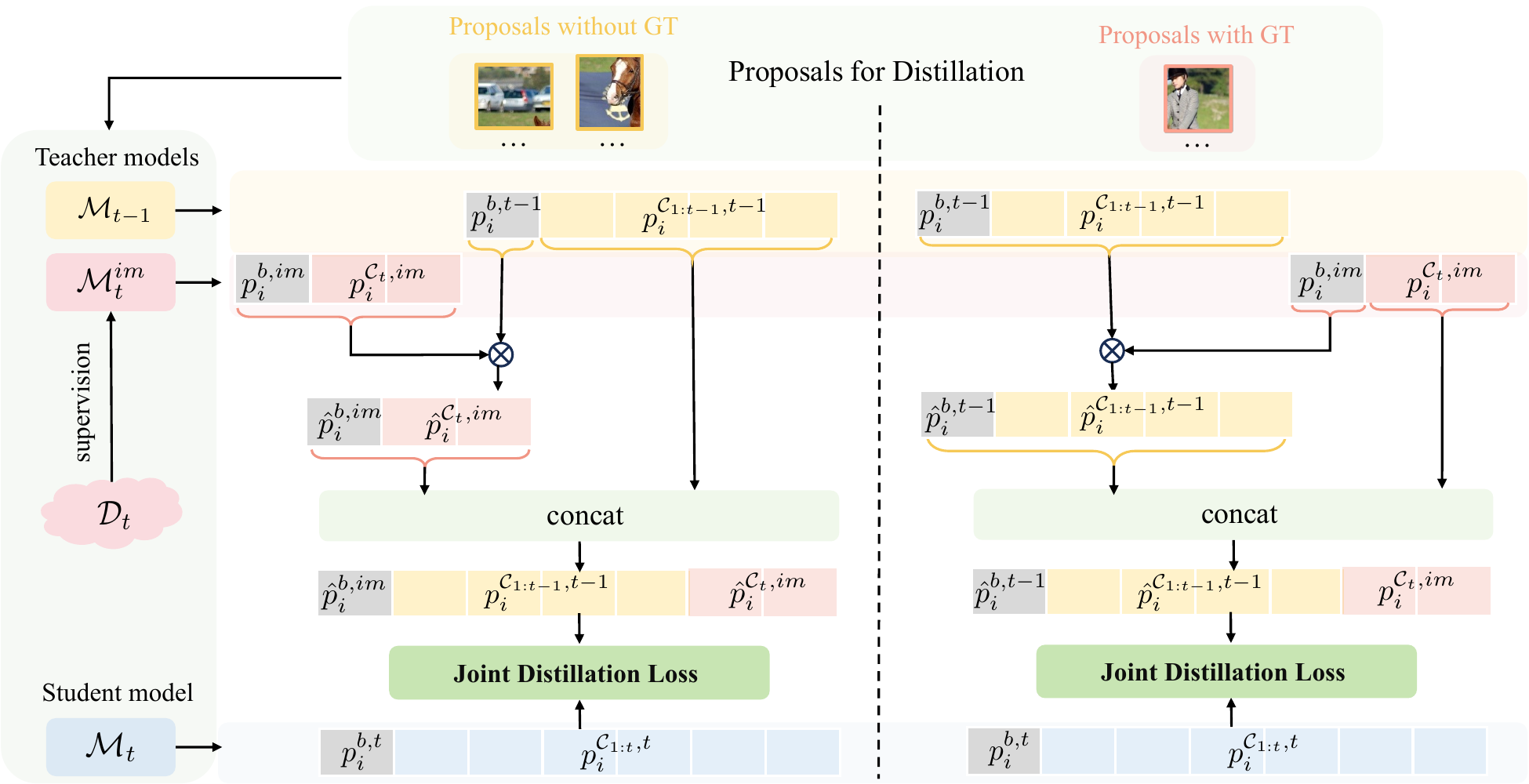}
    \caption{The illustration of Distillation with Future strategy. An intermediate teacher model trained on the current dataset is used to compensate for the lack of current class information in the old model. For proposals overlapping ground truth of the current stage, since the intermediate teacher is specialized in detecting current classes, we directly inherit its probabilities on current classes and use the old model to enrich its background probability with old class knowledge. On the contrary, for proposals that do not overlap with GT, the old model is preferred, and the intermediate model is used as compensation. Combining two teachers makes the distillation class by class.}
    \label{fig:dis-bpf}
\end{figure}

Distillation is an effective way to prevent forgetting in incremental object detection. However, distilling from the teacher model, which is biased toward old classes, inevitably hinders the learning for current classes. To address this challenge, we introduce the Distillation with Future (DwF) loss to distill from the teacher model in a more fine-grained and adaptive way. Different from MMA\cite{cermelli2022modeling}, which aligns the current classes and the background as a whole with the old model's background, we distill each class one by one, preserving the distinction between current classes and the background. However, the model $\mathcal{M}_{t-1}$ from the previous stage has not been trained on current classes, thus the distillation on current classes can not be performed. We introduce another intermediate teacher model $\mathcal{M}_{t}^{im}$, which is trained using the dataset from the current stage $\mathcal{D}_t$ in a fully supervised way, as a supplement. Taking the intermediate model $\mathcal{M}_{t}^{im}$ as a complementary teacher, we explicitly consider the current stage $t$ as the future stage of stage $t-1$, making the distillation future-aware.

As the old model $\mathcal{M}_{t-1}$ performs well on $\mathcal{C}_{1:t-1}$ while the intermediate model $\mathcal{M}_{t}^{im}$ is the expert in $\mathcal{C}_t$, we distill different regions with different combination of teacher models, as shown in \Cref{fig:dis-bpf}. Specifically, we divide the regions for distillation $\mathcal{R}$ into two subsets $\mathcal{R}_1,\mathcal{R}_2 \subset \mathcal{R}$ that based on their intersection over union with the ground truth labels for the new categories $\mathcal{C}_t$:
\begin{equation}
    \begin{aligned}
        \mathcal{R}_1=\{j\in\mathcal{R}:\forall i\in\mathcal{Y}_t,\mathrm{IoU}(b_j,b_i)\leq \lambda_2.\},\\
        \mathcal{R}_2=\{j\in\mathcal{R}:\forall i\in\mathcal{Y}_t,\mathrm{IoU}(b_j,b_i)> \lambda_2.\}.
    \end{aligned}
    \label{eq:lamda2}
\end{equation}

For regions $r_i \in \mathcal{R}_1$, which are likely to be the regions for old classes, we take $\mathcal{M}_{t-1}$ as the primary model for distillation and reconstruct its background representation with the model $\mathcal{M}_{t}^{im}$:
\begin{equation}
    \hat{p}_{i}^{c,im} = p_i^{c,im} \times p_i^{b,t-1}, \quad r_i \in \mathcal{R}_1,
\end{equation}
where $p_i^{b,t-1}$ and $p_i^{c,im}$ are the classification probabilities of background in the model $\mathcal{M}_{t-1}$ and the classification probabilities for current classes and background $c\in \mathcal{C}_t\cup\mathcal{B}$ in the model $\mathcal{M}_{t}^{im}$ for the region $r_i$ respectively. After weighting, $\sum\limits_{c=1}^{\mathcal{C}_t\cup\mathcal{B}}\hat{p}_{i}^{c,im} = p_i^{b,t-1}$. The final distillation probabilities for regions in $\mathcal{R}_1$ are $[p_{i}^{\mathcal{C}_{1:t-1},t-1}, \hat{p}_{i}^{\mathcal{C}_{t}\cup\mathcal{B},im}] \in \mathbb{R}^{|\mathcal{C}_{1:t}|+1}$.

On the contrary, for regions $r_i \in \mathcal{R}_2$, they are regions for current classes; thus, the intermediate model $\mathcal{M}_{t}^{im}$ is taken as the primary model for distillation, and its background representation is reconstructed by the model $\mathcal{M}_{t-1}$:
\begin{equation}
    \hat{p}_i^{c,t-1} = p_i^{c,t-1} \times p_i^{b,im}, \quad r_i \in \mathcal{R}_2,
\end{equation}
where $\sum\limits_{c=1}^{\mathcal{C}_{1:t-1}\cup\mathcal{B}} \hat{p}_i^{c,t-1} = p_i^{b,im}.$ The distillation probabilities for regions in $\mathcal{R}_2$ are $[\hat{p}_{i}^{\mathcal{C}_{1:t-1},t-1},$ $ p_{i}^{\mathcal{C}_{t},im}, \hat{p}_{i}^{\mathcal{B},t-1}] \in \mathbb{R}^{|\mathcal{C}_{1:t}|+1}$.

With the expanded probabilities from the complimentary teacher models, the distillation on the classification head can be performed using a conventional Kullback-Leibler divergence. Regarding the box distillation, we use the output boxes from the old model $\mathcal{M}_{t-1}$ for regions in $\mathcal{R}_1$ and the intermediate model $\mathcal{M}_{t}^{im}$ for $\mathcal{R}_2$. As a result, the complementary knowledge from two teachers, the expansion of background probability, and the combination of adaptive probability for different regions not only prevent the student from catastrophic forgetting but also facilitate the learning for current classes.

\section{Experiments}

\begin{table}[t]
  \centering
  \caption{mAP@0.5 results on single incremental step on PASCAL VOC 2007. The best performance in each is presented with \textbf{bold}, and the second best is presented with \underline{underlined}. Methods with $*$ use exemplars.}
  \resizebox{1.0\linewidth}{!}{
  \begin{tabular}{l || c c c c | c c c c | c c c c | c c c c}
    \toprule
    \multirow{2}{*}{\textbf{Method}}     & \multicolumn{4}{c|}{$\textbf{19-1}$} & \multicolumn{4}{c|}{$\textbf{15-5}$} & \multicolumn{4}{c|}{$\textbf{10-10}$} & \multicolumn{4}{c}{$\textbf{5-15}$}\\
     & $\textbf{1-19}$ & $\textbf{20}$ &\cellcolor{Gray}$\textbf{1-20}$& \textbf{Avg} &$\textbf{1-15}$ & $\textbf{16-20}$ &\cellcolor{Gray}$\textbf{1-20}$ & \textbf{Avg} & $\textbf{1-10}$ & $\textbf{11-20}$ &\cellcolor{Gray}$\textbf{1-20}$& \textbf{Avg} & $\textbf{1-5}$ & $\textbf{5-15}$ &\cellcolor{Gray}$\textbf{1-20}$& \textbf{Avg}\\
    \midrule
    \midrule
    Joint Training & 76.0 & 76.7 & \cellcolor{Gray}76.1& 76.4 &78.0 &70.4 &\cellcolor{Gray}76.1& 74.2 & 75.9 &76.3 &\cellcolor{Gray}76.1& 76.1 &72.4 &77.3 &\cellcolor{Gray}76.1& 74.9\\
    Fine-tuning & 12.0 & 62.8 & \cellcolor{Gray}14.5& 37.4 & 14.2 & 59.2 & \cellcolor{Gray}25.4& 36.7 & 9.5 & 62.5 & \cellcolor{Gray}36.0& 36.0 & 6.9 & 63.1 & \cellcolor{Gray}49.1& 35.0 \\
    \midrule
    ORE*\cite{joseph2021towards} &69.4 &60.1 &\cellcolor{Gray}68.9&64.7 &71.8 &58.7 &\cellcolor{Gray}68.5&65.2 &60.4 &68.8 &\cellcolor{Gray}64.6&64.6 &- &- &\cellcolor{Gray}-&-\\
    OW-DETR*\cite{gupta2022ow} &70.2 &62.0 &\cellcolor{Gray}69.8&66.1 &72.2 &59.8 &\cellcolor{Gray}69.1&66.0 &63.5 &67.9 &\cellcolor{Gray}65.7&65.7 &- &- &\cellcolor{Gray}-&-\\
    ILOD-Meta*\cite{joseph2021incremental} &70.9 &57.6 &\cellcolor{Gray}70.2&64.2 &71.7 &55.9 &\cellcolor{Gray}67.8&63.8 &68.4 &64.3 &\cellcolor{Gray}66.3&66.3 &- &- &\cellcolor{Gray}-&-\\
    ABR*\cite{liu2023augmented} & 71.0 &\textbf{69.7} &\cellcolor{Gray}70.9&\textbf{70.4} &73.0 &\textbf{65.1} &\cellcolor{Gray}\underline{71.0}&\underline{69.1} &\underline{71.2} &\underline{72.8} &\cellcolor{Gray}\underline{72.0}&\underline{72.0} & 64.7 & \underline{71.0} & \cellcolor{Gray}\underline{69.4}& \underline{67.9}\\
    \midrule
    Faster ILOD\cite{ren2015faster} &68.9 &61.1 &\cellcolor{Gray}68.5&65.0 &71.6 &56.9 &\cellcolor{Gray}67.9&64.3 &69.8 &54.5 &\cellcolor{Gray}62.1&62.1 & 62.0 & 37.1 & \cellcolor{Gray}43.3& 49.6\\
    PPAS\cite{zhou2020lifelong} & 70.5 &53.0 &\cellcolor{Gray}69.2&61.8 &- &- &\cellcolor{Gray}-&- &63.5 &60.0 &\cellcolor{Gray}61.8&61.8 &- &- &\cellcolor{Gray}-&-\\
    MVC\cite{yang2022multi} &70.2 &60.6 &\cellcolor{Gray}69.7&65.4 &69.4 &57.9 &\cellcolor{Gray}66.5&63.7 &66.2 &66.0 &\cellcolor{Gray}66.1&66.1 &- &- &\cellcolor{Gray}-&-\\
    PROB\cite{zohar2023prob} & \underline{73.9} & 48.5 &\cellcolor{Gray}\underline{72.6}& 61.5 &\underline{73.5} & 60.8 &\cellcolor{Gray}70.1& 67.0 & 66.0 & 67.2 & \cellcolor{Gray}66.5 & 66.5 &- &- &\cellcolor{Gray}-&-\\
    PseudoRM\cite{yang2023pseudo} & 72.9 & \underline{67.3} & \cellcolor{Gray}\underline{72.6} & \underline{70.1} & 73.4 & 60.9 &\cellcolor{Gray}70.3& 66.9 & 69.1 & 68.6 &\cellcolor{Gray}68.9& 68.9 &- &- &\cellcolor{Gray}-&-\\
    MMA\cite{cermelli2022modeling} & 71.1 & 63.4 & \cellcolor{Gray}70.7& 67.2 &73.0 &60.5 &\cellcolor{Gray}69.9&66.7 &69.3 &63.9 &\cellcolor{Gray}66.6&66.6 & \textbf{66.8} & 57.2 & \cellcolor{Gray}59.6& 62.0\\
    \textbf{BPF (Ours)} & \textbf{74.5} & 65.3 &\cellcolor{Gray}\textbf{74.1}& 69.9 &\textbf{75.9} &\underline{63.0} & \cellcolor{Gray}\textbf{72.7}& \textbf{69.5} &\textbf{71.7} &\textbf{74.0} &\cellcolor{Gray}\textbf{72.9}& \textbf{72.9} & \underline{66.4} & \textbf{75.3} & \cellcolor{Gray}\textbf{73.0}& \textbf{70.9}\\
    \bottomrule
  \end{tabular}}
  \label{tab:voc-single}
\end{table}

\subsection{Experiment Settings}

\noindent\textbf{Datasets and Evaluation Metrics.}
Following previous works\cite{cermelli2022modeling,gupta2022ow,joseph2021incremental,joseph2021towards,zhou2020lifelong,peng2020faster,liu2023augmented,yang2022multi,shmelkov2017incremental}, we evaluate our method on PASCAL VOC 2007\cite{everingham2010pascal} and MS COCO 2017\cite{lin2014microsoft} datasets.
PASCAL VOC 2007 dataset comprises 9,963 images across 20 categories.
The COCO 2017 dataset encompasses objects from 80 categories, with around 118k images for training and 5,000 images for validation.
The mean average precision at the 0.5 IoU threshold (mAP@0.5) is used as the primary evaluation metric for the VOC dataset, and the mean average precision ranging from 0.5 to 0.95 is the main evaluation metric for the COCO dataset.

For each incremental setting (A-B), the first number A denotes the number of classes in the first stage and the second number is the number of classes newly introduced in each new stage. Note that the columns with gray background in the table of experimental results represent the average AP among all classes.

\vspace{0.5em}
\noindent\textbf{Implementation Details.}
Similar to \cite{cermelli2022modeling,joseph2021incremental,joseph2021towards,liu2023augmented,peng2020faster}, we build our incremental object detector based on Faster R-CNN\cite{ren2015faster} with R50. Our method can easily be adapted to transformer-based detectors~\cite{detr, deformabledetr, adamixer}. We conduct the experiments under a strict \textbf{rehearsal-free} setting, where no memory is used.
We set $\eta=0.75$, $\lambda_1=0.7$, $\lambda_2=0.5$. For $\mathcal{W}$ in \Cref{eq:pseudo-labels}, we use an IOU threshold of 0.3 to divide it into two sets. For the set with $\mathrm{IoU}<0.3$, the supervision signal weights are set to 1.0, while the other set is assigned a weight of 0.3.

\subsection{Quantitative Evaluation}
Following previous work\cite{cermelli2022modeling,liu2023augmented,chen2019new,peng2020faster,shmelkov2017incremental,yang2022multi,zhou2020lifelong}, our method is evaluated on settings with a range of initial classes and incorporating one or more incremental tasks. We benchmark our method against two baselines: Fine-Tuning, where the model is incrementally trained on new data without any regularization strategy or data replay, and Joint Training, which involves training the model on the complete dataset using all annotations. 

\begin{table}[t]
  \centering
  \setlength{\tabcolsep}{4pt}
  \caption{mAP@0.5 results on multiple incremental steps on PASCAL VOC 2007. The best performance in each is presented with \textbf{bold}, and the second best is presented with \underline{underlined}. Methods with $*$ use exemplars.}
  \resizebox{1.0\linewidth}{!}{
  \begin{tabular}{l || c c c | c c c | c c c | c c c | c c c}
    \toprule
    \multirow{2}{*}{\textbf{Method}}     & \multicolumn{3}{c|}{\textbf{10-5 (3 tasks)}} & \multicolumn{3}{c|}{\textbf{5-5 (4 tasks)}} & \multicolumn{3}{c|}{\textbf{10-2 (6 tasks)}} & \multicolumn{3}{c|}{\textbf{15-1 (6 tasks)}} & \multicolumn{3}{c}{\textbf{10-1 (10 tasks)}}\\
     & \textbf{1-10} & \textbf{11-20} & \cellcolor{Gray}\textbf{1-20} & \textbf{1-5} & \textbf{6-20} & \cellcolor{Gray}\textbf{1-20} & \textbf{1-10} & \textbf{11-20} & \cellcolor{Gray}\textbf{1-20} & \textbf{1-15} & \textbf{16-20} & \cellcolor{Gray}\textbf{1-20}& \textbf{1-10} & \textbf{11-20} & \cellcolor{Gray}\textbf{1-20}\\
    \midrule
    \midrule
    Joint Training &75.9 &76.3 &\cellcolor{Gray}76.1 &72.4 &77.3 &\cellcolor{Gray}76.1 &75.9 &76.3 &\cellcolor{Gray}76.1 &78.0 &70.4 &\cellcolor{Gray}76.1&75.9 &76.3 &\cellcolor{Gray}76.1\\
    Fine-tuning\cite{liu2023augmented} & 5.3 &30.6 &\cellcolor{Gray}18.0 &0.5 &18.3 &\cellcolor{Gray}13.8 &3.8 &13.6 &\cellcolor{Gray}8.7 &0.0 &10.5 &\cellcolor{Gray}5.3 &0.0 &5.1 &\cellcolor{Gray}2.6\\
    \midrule
    ABR*\cite{liu2023augmented} & \underline{68.7} &\underline{67.1} &\cellcolor{Gray}\underline{67.9} &\textbf{64.7} &\underline{56.4} &\cellcolor{Gray}\underline{58.4} &\underline{67.0} &\textbf{58.1} &\cellcolor{Gray}\textbf{62.6} &\underline{68.7} &\textbf{56.7} &\cellcolor{Gray}\underline{65.7} & 62.0 & \textbf{55.7} & \cellcolor{Gray}\textbf{58.9}\\
    \midrule
    Faster ILOD\cite{ren2015faster} &68.3 &57.9 &\cellcolor{Gray}63.1 &55.7 &16.0 &\cellcolor{Gray}25.9 &64.2 &48.6 &\cellcolor{Gray}56.4 &66.9 &44.5 &\cellcolor{Gray}61.3 &52.9 &41.5 &\cellcolor{Gray}47.2\\
    MMA\cite{cermelli2022modeling} &66.7 &61.8 &\cellcolor{Gray}64.2 &\underline{62.3} &31.2 &\cellcolor{Gray}38.9 &65.0 &53.1 &\cellcolor{Gray}59.1 &68.3 &\underline{54.3} &\cellcolor{Gray}64.1 &59.2 &\underline{48.3} &\cellcolor{Gray}53.8\\
    \textbf{BPF (Ours)} & \textbf{69.1} & \textbf{68.2} &\cellcolor{Gray}\textbf{68.7} &60.6 &\textbf{63.1} &\cellcolor{Gray}\textbf{62.5} & \textbf{68.7}& \underline{56.3}&\cellcolor{Gray}\underline{62.5}& \textbf{71.5}& 53.1 & \cellcolor{Gray}\textbf{66.9} &\textbf{62.2} &\underline{48.3} & \cellcolor{Gray}\underline{55.2}\\
    \bottomrule
  \end{tabular}}
  \vspace{-3.0em}
  \label{tab:voc-multi}
\end{table}

\begin{minipage}[l]{0.5\linewidth}
    \centering
    \footnotesize
    \captionof{table}{mAP results on COCO2017. Methods with $*$ use exemplars.}
    \resizebox{1.0\linewidth}{!}{
    \begin{tabular}{l || c c c| c c c}
    \toprule
    \multirow{2}{*}{\textbf{Method}}     & \multicolumn{3}{c|}{ $\textbf{40-40}$} & \multicolumn{3}{c}{$\textbf{70-10}$}\\
     & $AP$ & $AP_{50}$ & $AP_{75}$ & $AP$ & $AP_{50}$ & $AP_{75}$\\
    \midrule
    \midrule
    Joint Training & 36.7 & 57.8 & 39.8 & 36.7 & 57.8 & 39.8 \\
    Fine-tuning\cite{liu2023augmented} & 19.0 &31.2 & 20.4 & 5.6 & 8.6 & 6.2\\
    \midrule
    ILOD-Meta*\cite{joseph2021incremental} & 23.8 & 40.5 & 24.4 & - &  -& -\\
    ABR*\cite{liu2023augmented} & \textbf{34.5} & \textbf{57.8} & \underline{35.2} &\underline{31.1} &\underline{52.9} &\underline{32.7}\\
    \midrule
    Faster ILOD\cite{peng2020faster} &20.6 &40.1 &- &21.3 &39.9 &-\\
    PseudoRM\cite{yang2023pseudo} & 25.3 & 44.4 & - & - & -& -\\
    MMA\cite{cermelli2022modeling} & 33.0 & \underline{56.6} & 34.6 & 30.2 &52.1 & 31.5\\
    \textbf{BPF (Ours)} & \underline{34.4} & 54.3 & \textbf{37.3} & \textbf{36.2} & \textbf{56.8} & \textbf{38.9}\\
    \bottomrule
  \end{tabular}}
  \label{tab:result-coco}
\end{minipage}
\hspace{0.2em}
\begin{minipage}[c]{0.4\linewidth}
    \begin{minipage}[c]{1\linewidth}
        \centering
        \footnotesize
        \captionof{table}{Ablation study of various combinations of teacher models.}
        \resizebox{0.85\linewidth}{!}{
        \begin{tabular}{c c| c c c }
        \toprule
        \multicolumn{2}{c|}{\textbf{Distillation}}   &   \multicolumn{3}{c}{\textbf{VOC(10-10})}\\
        $\mathcal{L}_{dist,cls}^{roi}$ & $\mathcal{L}_{dist,bbox}^{roi}$ & \textbf{1-10} & \textbf{11-20} & \cellcolor{Gray}\textbf{1-20}\\
        \midrule
        \midrule
        $\lambda_2=1.0 $  & part boxes & 71.5 & 73.3 & \cellcolor{Gray}72.4      \\
        $\lambda_2=0.5 $  & part boxes & \textbf{71.7} & 74.0 & \cellcolor{Gray}\textbf{72.9}\\
        $\lambda_2=0.5 $  & all boxes  & 71.3 & \textbf{74.4} & \cellcolor{Gray}\textbf{72.9}\\
        \bottomrule
      \end{tabular}
        }
        \label{tab:abl-dis}
    \end{minipage}

    \begin{minipage}[c]{1\linewidth}
      \centering
      \captionof{table}{Effect of BF.}
      \scalebox{0.5}{
      \begin{tabular}{c|c c c|ccc|ccc}
        \toprule
        \multirow{2}{*}{\textbf{BF}} & \multicolumn{3}{c|}{\textbf{VOC(5-15)}}& \multicolumn{3}{c|}{\textbf{VOC(10-10)}}& \multicolumn{3}{c}{\textbf{VOC(15-5)}}\\
        & \textbf{1-5} & \cellcolor{Gray}\textbf{6-20} & \textbf{1-20} & \textbf{1-10} & \cellcolor{Gray}\textbf{11-20} &\textbf{1-20} &\textbf{1-15} & \cellcolor{Gray}\textbf{16-20} &\textbf{1-20}\\
        \midrule
        $\times$& 66.3 &\cellcolor{Gray}74.4 &72.4 &71.2 &\cellcolor{Gray}73.3 &72.3 &75.6 &\cellcolor{Gray}62.8 &72.4\\
        $\checkmark$ &\textbf{66.4} &\cellcolor{Gray}\textbf{75.3} &\textbf{73.0} &\textbf{71.7} &\cellcolor{Gray}\textbf{74.0} &\textbf{72.9} &\textbf{75.9} &\cellcolor{Gray}\textbf{63.0} &\textbf{72.7}\\
        \bottomrule
      \end{tabular}}
      \label{tab:abl_bpf}
    \end{minipage}
\end{minipage}

\vspace{0.5em}
\noindent\textbf{PASCAL VOC 2007.}
For PASCAL VOC 2007, we order the classes alphabetically and evaluate our method with one or multiple training steps. We perform our experiments by adding 1 (19-1), 5 (15-5), 10 (10-10), or 15 (5-15) classes in a single incremental step. 
For multi-step incremental settings, we evaluate 10-5, 5-5, 10-2, 15-1, and 10-1 settings, where we add 5, 5, 2, 1, and 1 classes respectively at every step until all 20 classes are seen.

\noindent\textbf{- Single-step Incremental Settings:}
\Cref{tab:voc-single} shows our BPF methods against the existing methods using rehearsal or not. 
Rehearsal-based methods are not compared fairly with our BPF since we do not store old samples and use replay memory.
As shown in \Cref{tab:voc-single}, BPF consistently outperforms all previous methods, including those designed to combat forgetting using exemplars, validating the superiority of our approach.
In particular, in the 19-1, 15-5, 10-10, and 5-15 settings, BPF significantly improved over MMA\cite{cermelli2022modeling} by 3.4\%, 2.8\%, 6.3\%, and 13.4\% on mAP@0.5 across all classes.
Similarly, BPF outperforms the best rehearsal-based method ABR\cite{liu2023augmented} by 3.2\%, 1.7\%, 0.9\%, and 3.6\%. The \textit{Avg} metric equally averages old and new classes mAP, which straightly reports the incremental ability without the influence of the number of classes. BPF also outperforms most methods on the \textit{Avg} metric, demonstrating BPF's adaptiveness in learning new classes and preserving the knowledge of old classes.

\noindent\textbf{- Multi-step Incremental Settings:} The issues of inconsistent optimization objectives across multiple stages and catastrophic forgetting are more crucial under the longer incremental settings.
As shown in \Cref{tab:voc-multi}, BPF consistently outperforms MMA\cite{cermelli2022modeling} across all the settings. Specifically, BPF improves over MMA by 1.4\%, 2.8\%, 3.4\%, 4.5\%, and even 23.6\% at \textbf{1-20} mAP@0.5 under the 10-1, 15-1, 10-2, 10-5, and 5-5 settings and enjoy improvement across all learning stages.
Moreover, we find that even without storing memory, BPF still outperforms ABR~\cite{liu2023augmented} by 0.8\% and 4.1\% on overall mAP@0.5, 1.1\% and 6.7\% mAP@0.5 on new classes under 10-5 and 5-5 settings. In the settings of 10-2 and 10-1, limited by the small incremental data, BPF is inferior to ABR, but considering we do not require memory, these losses are acceptable.

\vspace{0.5em}
\noindent\textbf{MS COCO 2017.}
On the COCO2017 dataset, we perform experiments on 40-40 and 70-10 settings, adding 40 and 10 classes, respectively, following \cite{liu2023augmented}.
As illustrated in \Cref{tab:result-coco}, 
our method improves over MMA on average $AP$ by 1.4\% on 40-40 settings and by 6.0\% on 70-10 settings. These results once again confirm the effectiveness of our method.

\subsection{Analysis and Ablation Study}
We examine the contributions of the ``Bridge the Past'', ``Bridge the Future'', and ``Distillation with Future'' in \Cref{tab:abl_overall} within the VOC 10-10 and 10-5 settings. We take the unbiased knowledge distillation proposed by \cite{cermelli2022modeling} as the baseline model.
By bridging the past, our model (b) aligns its optimization objectives with earlier ones, effectively reducing catastrophic forgetting of old classes, greatly enhancing the performance on detecting old classes. Compared to the baseline (a), it significantly improved by 13.1\% in the old classes and by 6.4\% for all classes on the 10-10 setting.
Owing to bridging the future, our model (c) maintains consistent optimization objectives with future models regarding the background, making it easier to incrementally learn new classes. Compared to the baseline (a) on the 10-10 setting, it improves by 2.9\% and 0.9\% on the old classes and the new classes, respectively. Combining Past and Future (d), the model outperforms consistently on each stage. In the Distillation with Future strategy, we leverage the intermediate model to aid the old model in modeling new classes, conducting distillation across all categories for the current model. \Cref{tab:abl_overall} (e) shows that a comprehensive joint teacher model facilitates improved learning of all categories through knowledge distillation, with a significant improvement (+1.3\% AP) in new classes. Similar results can be found in the 10-5 setting.

\begin{table}[t]
  \centering
  \setlength{\tabcolsep}{4pt}
  \caption{Ablation study on each component.}
  \scalebox{0.8}{
  \begin{tabular}{c|c | c | c | c c c | c c c c}
    \toprule
    \multirow{2}{*}{\textbf{Model}} & \textbf{Bridge}  & \textbf{Bridge} & \textbf{Distillation} &  \multicolumn{3}{c|}{\textbf{VOC(10-10)}} & \multicolumn{4}{c}{\textbf{VOC(10-5)}}\\
    & \textbf{the Past} & \textbf{the future} & \textbf{with Future} & \textbf{1-10} & \textbf{11-20} & \cellcolor{Gray}\textbf{1-20}& \textbf{1-10} & \textbf{11-15} & \textbf{16-20} & \cellcolor{Gray}\textbf{1-20}\\
    \midrule
    \midrule
     (a) &               &               &               & 58.1 & 72.4 & \cellcolor{Gray}65.3 & 54.4 & 69.9 & 59.3 & \cellcolor{Gray}59.5\\
     (b) &$\checkmark$   &               &                & 71.2 & 72.1 & \cellcolor{Gray}71.7 & 69.1 & 73.5 & 60.0 & \cellcolor{Gray}67.9\\
     (c) &               & $\checkmark$  &               & 61.0 & 73.3 & \cellcolor{Gray}67.1 & 54.8 & 70.7& 58.7 & \cellcolor{Gray}59.8\\   
     (d) &$\checkmark$   & $\checkmark$  &               & \textbf{71.9} & 72.7 & \cellcolor{Gray}72.3 & \textbf{70.4} & 73.7 & 58.7 & \cellcolor{Gray}68.3\\
     (e) &$\checkmark$   & $\checkmark$  & $\checkmark$  & 71.7 & \textbf{74.0} & \cellcolor{Gray}\textbf{72.9} & 69.1 & \textbf{75.2} & \textbf{61.2} & \cellcolor{Gray}\textbf{68.7}\\
    \bottomrule
  \end{tabular}}
  \label{tab:abl_overall}
\end{table}

We further verify the ``Bridge the future'' (BF) in \Cref{tab:abl_bpf}.
There is a clear trend that as the number of considered future classes increases, the BF shows increasing improvement in future classes (+0.9 \% in 5-15) without degrading the old classes' performance.

We also conducted a quantitative analysis of Bridge Past and Future. Under the second stage of the VOC 10-5-5 (3 tasks), the original background boxes have a 92.6\% and 83.9\% Recall50 rate for old and future classes, while the Recall50 of pseudo labels for old classes and discarded boxes for future classes are 66.1\% and 17.6\% respectively, demonstrating the necessity and effectiveness of Bridge the Past and Bridge the Future.

\Cref{tab:abl-dis} presents experiments on the effect of different combinations of teacher models.
$\lambda_2$ in \Cref{eq:lamda2} determines using which teacher model as the primary. Using the intermediate expert model as the primary on regions overlapping with gt ($\lambda_2 = 0.5$) outperforms using the old model ($\lambda_2 = 1.0$), showing that distilling new class objects with the old model hinders their learning. For box distillation, we find that distilling boxes on primary classes (part boxes, \ie, only boxes for old classes in $\mathcal{R}_1$ and boxes for new classes in $\mathcal{R}_2$ participate in distillation while others are ignored) performs similarly with all classes (all boxes).

\begin{figure}[t]
    \centering
    \includegraphics[width=1.0\linewidth]{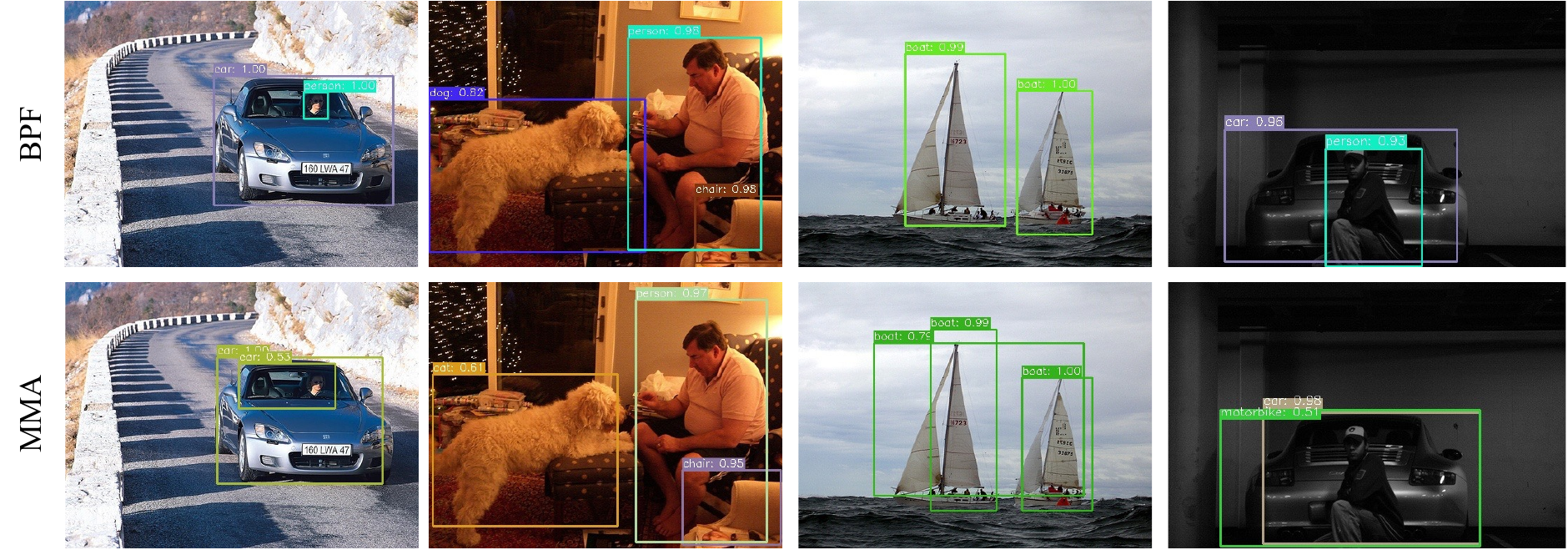}
    \caption{Qualitative results for the model trained under the 10+10 setting on the VOC 2007 test set. `boat', `cat', `chair', and `car' are old classes from the first stage, and `person' and `dog' are the classes from the second stage. Compared with MMA (bottom row), our BPF (top row) can produce reliable predictions on both old and new classes.}
    \label{fig:vis-comparsion}
\end{figure}

\subsection{Visualization}
\noindent\textbf{Visualization of Detection Results.} We visualize the detection results in \Cref{fig:vis-comparsion} to illustrate the significant improvement compared to the previous methods qualitatively. Our method accurately detects both new and old class objects simultaneously. While MMA fails to detect old classes accurately, suffering from catastrophic forgetting.

\begin{figure}[t]
    \centering
    \includegraphics[width=1.0\linewidth]{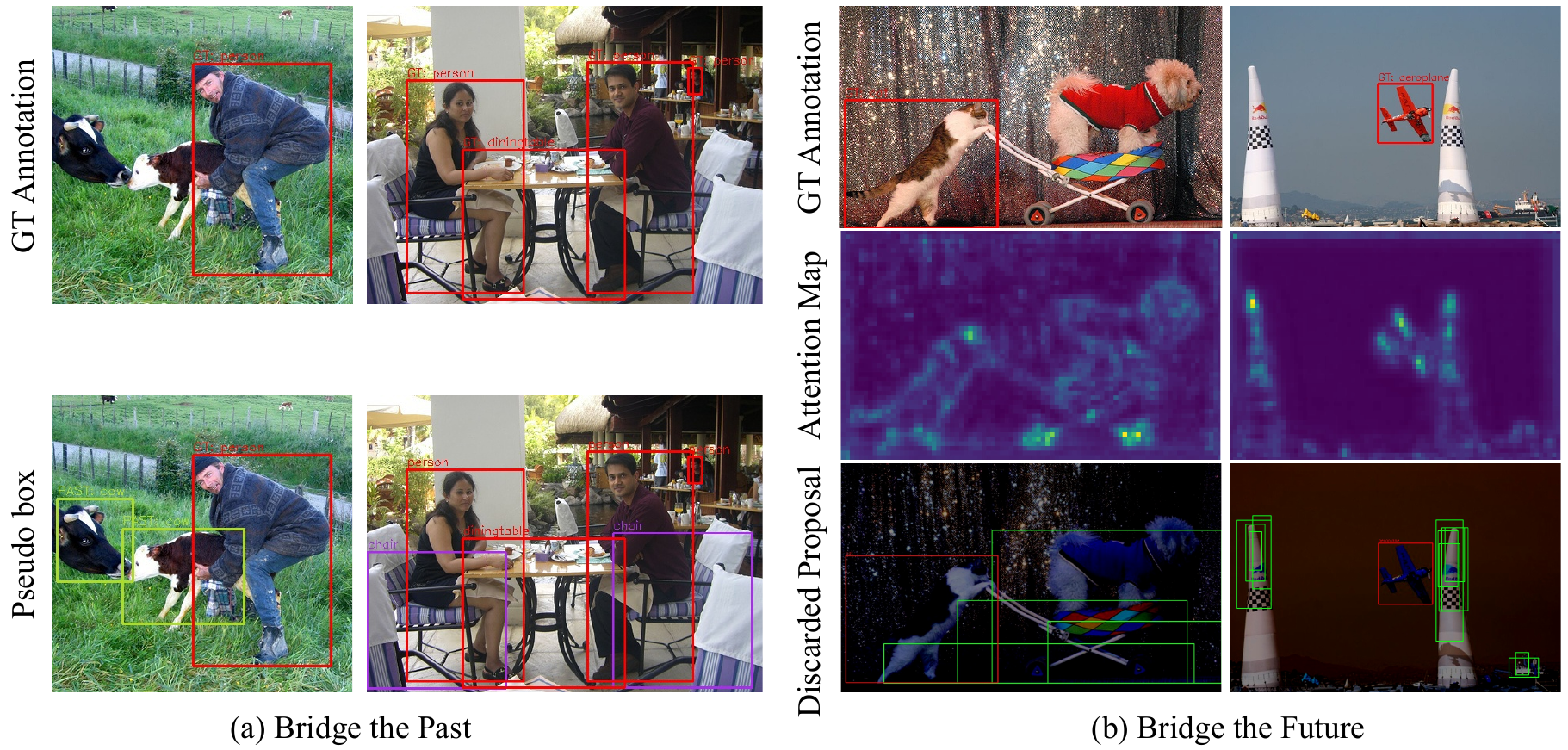}
    \caption{Visualization of Bridge Past and Future. Boxes in \textcolor{red}{red} represent the ground truth in the current stage. \textbf{(a)} In Bridge the Past, we effectively constructed pseudo labels of past classes. \textbf{(b)} In Bridge the Future, salient objects (marked in \textcolor{green}{green} boxes) can be easily detected from the attention maps and are excluded from the background regions. Best viewed in color.}
    \label{fig:vis-method}
\end{figure}

\vspace{0.5em}
\noindent\textbf{Visualization of Bridge Past and Future.} We visualized the module of Bridge the Past and Bridge the Future separately in \Cref{fig:vis-method}. We effectively model the missing annotations for old class objects to bridge the past. As demonstrated in \Cref{fig:vis-method}(b), the attention maps clearly differentiate the foreground from the background, regardless of the presence of annotations. The discarded boxes (the third row) validate the effectiveness of our method.

\section{Conclusions}
\label{sec:conclusion}
\noindent\textbf{Limitations.}
In our Bridge the Past procedure, we assume several objects of old classes may appear in the current training data. However, when the number of incremental classes is limited, \eg, increasing a single class in each stage, objects of old classes may rarely occur due to limited training images. This is expected to be alleviated by generating samples of old classes via the copy-paste strategy as in ABR\cite{liu2023augmented}, while it may introduce little stored samples. Detailed discussion can be found in the Supplement.

\vspace{0.4em}
\noindent\textbf{Conclusions.}
In this work, we find that the concurrence of classes from different learning stages causes a severe information asymmetry, not only causing catastrophic forgetting for old classes but also hindering the learning of new classes. To tackle the problem, we propose the Bridge Past and Future method, which uses pseudo labels from the old model to fill in the missing annotations and exclude some potential future objects from the background, keeping the learning consistent across all stages. Further, Distillation with Future loss is proposed to solve the problem of the old teacher model's lack of knowledge of new classes. By combining the knowledge from the past to the future, our method consistently outperforms others across different learning stages in most incremental settings. To the best of our knowledge, we are the first to consider future classes during the learning, shedding light on a new aspect of Incremental Object Detection.

\vspace{0.4em}
\noindent\textbf{Acknowledgments.} This work was supported partially by the National Key Research and Development Program of China (2023YFA1008503), NSFC(U21A20 471), Guangdong NSF Project (No. 2023B1515040025, 2020B1515120085).

\bibliographystyle{splncs04}
\bibliography{main}

\newpage
\input{arxiv_supp}

\end{document}

%% file: arxiv_supp.tex
\def\Ie{\emph{I.e}\onedot}
\def\cf{\emph{c.f}\onedot} \def\Cf{\emph{C.f}\onedot}
\def\etc{\emph{etc}\onedot} \def\vs{\emph{vs}\onedot}
\def\wrt{w.r.t\onedot} 
\def\aka{a.k.a\onedot} 
\def\dof{d.o.f\onedot}
\def\etal{\emph{et al}\onedot}
\def\ie{\textit{i.e.}}
\def\eg{\textit{e.g.}}
\renewcommand\thetable{S\arabic{table}}
\renewcommand\thefigure{S\arabic{figure}}
\renewcommand\thesection{S\arabic{section}}

\title{\textit{Supplementary Materials for}\\Bridge Past and Future: Overcoming Information Asymmetry in Incremental Object Detection}

\titlerunning{ }
\author{ }
\authorrunning{ }
\institute{ }
\maketitle

In this supplementary material, we present the detailed results for the 5-5 multi-step incremental setting (\Cref{sec:s1}), more discussion on possible limitations (\Cref{sec:s2}), and some ablation studies on selection metrics for potential future class objects in Bridge the Future (\Cref{sec:s33}). We also provide a detailed process of Distillation with Future (\Cref{sec:s3}) and more analysis on the necessity for Distillation with Future (\Cref{sec:s4}). 

\section{Detailed Results for the Multi-Step Incremental Setting}
\label{sec:s1}

\Cref{tab:15-1_detailed} presents the results of our experiments under 5-5 multi-step incremental setting on the PASCAL-VOC 2007 dataset\cite{everingham2010pascal}.
We simulate this scenario by training the detector on images from the first 5 classes and then adding the $6^{th}\sim 20^{th}$ classes five by five. The table shows the class-wise average precision (AP@0.5) and the corresponding mean average precision (mAP@0.5). Our proposed BPF method outperforms the previous state-of-the-art rehearsal-based method ABR\cite{liu2023augmented}.

\begin{table}[ht]
  \centering
  \setlength{\tabcolsep}{4pt}
  \caption{Per-Class AP@0.5 and Overall mAP@0.5 values in different task on PASCAL-VOC 2007 5-5 setting.}
  \vspace{-0.2em}
  \resizebox{1.0\linewidth}{!}{
  \begin{tabular}{l|l | c c c c c c c c c c c c c c c c c c c c c c c c}
    \toprule
    Class Split & Method & aero &cycle &bird &boat &bottle &bus &car &cat &chair &cow &\cellcolor{Q-Gray}mAP-task1 &table &dog &horse &bike &person &\cellcolor{Q-Gray}mAP-task2 &plant &sheep &sofa &train &tv &\cellcolor{Q-Gray}mAP-task3 &\cellcolor{Gray}mAP-total\\
    \hline
    1-20 &Joint Training &76.4 & 84.7 & 77.1 & 62.7 & 61.3 & 82.2 & 87.5 & 85.9 & 57.8 & 83.0 &\cellcolor{Q-Gray}75.9 &70.5 & 84.4 & 86.7 & 83.7 & 85.6 &\cellcolor{Q-Gray}82.2 &46.7 & 77.7 & 70.7 & 80.3 & 76.7 &\cellcolor{Q-Gray}70.4 &\cellcolor{Gray}76.1\\
    \hline
    (1-5)+ &ABR\cite{liu2023augmented} & 71.7 &82.6 &69.5 &53.6 &63.8 &63.0 &79.0 &68.5 &47.0 &78.4 &\cellcolor{Q-Gray} 67.7 & & & & & &\cellcolor{Q-Gray} & & & & & &\cellcolor{Q-Gray} &\cellcolor{Gray}67.7\\
    6-10 & \textbf{BPF (Ours)} & 70.1 & 78.3 &68.3 &55.9 &60.1 &64.9 &81.3 &73.3 &52.3 &79.8 &\cellcolor{Q-Gray}\textbf{68.4} & & & & & &\cellcolor{Q-Gray} & & & & & &\cellcolor{Q-Gray} &\cellcolor{Gray}\textbf{68.4}\\ 
    \hline
    (1-10)+ &ABR\cite{liu2023augmented} & 68.5 &79.6 &67.3 &51.9 &56.7 &60.2 &75.2 &62.8 &38.6 &62.0 &\cellcolor{Q-Gray}62.3 &54.0 &66.3 &76.9 &74.5 &77.3 &\cellcolor{Q-Gray} 69.8& & & & & &\cellcolor{Q-Gray} &\cellcolor{Gray}64.8\\
    11-15 & \textbf{BPF (Ours)} & 69.2 &79.8 &64.9 &54.5 &56.4 &65.3 &80.6 &68.8 &48.3 &71.9 &\cellcolor{Q-Gray}\textbf{66.0}  &58.6 &68.8 &77.5 &74.5 &80.6 &\cellcolor{Q-Gray} \textbf{72.0} & & & & & &\cellcolor{Q-Gray} & \cellcolor{Gray}\textbf{68.0}\\
    \hline
    (1-15)+ &ABR\cite{liu2023augmented} & 69.3 &80.0 &65.6 &53.9 &54.6 &52.2 &75.5 &69.4 &34.3 &69.6 &\cellcolor{Q-Gray}\textbf{62.4} &22.9 &41.8 &48.7 &53.7 &60.8 &\cellcolor{Q-Gray} 45.6 &39.6 &71.3 &59.2 &76.1 &70.4 &\cellcolor{Q-Gray}\textbf{63.3} &\cellcolor{Gray}58.4\\
    16-20 & \textbf{BPF (Ours)}& 64.8 &75.9 &56.7 &51.8 &53.8 &51.1 & 78.8 &63.6 &47.1 &63.5 &\cellcolor{Q-Gray}60.7 &56.2 &64.1 &75.9 &71.2 &79.4 &\cellcolor{Q-Gray} \textbf{69.4} &40.4 &61.3 &61.7 &61.8 &70.5 &\cellcolor{Q-Gray}59.1 &\cellcolor{Gray}\textbf{62.5}\\
    \bottomrule
  \end{tabular}}
  \vspace{-0.5em}
  \label{tab:15-1_detailed}
\end{table}

\section{Discussion on Possible Limitations}
\label{sec:s2}
As discussed in Section \textcolor{red}{5} in the main text, our method is based on a prior that the old classes always appear in the new training dataset so that they can be utilized to prevent catastrophic forgetting. However, when the newly introduced dataset is too small to find sufficient old class objects, especially in incremental scenarios that only add a few classes at each stage, our Bridge the Past method cannot be performed effectively, which hinders both old and new classes performance as the balance of old and new classes is a trade-off.

However, the limitation can be simply addressed by using some memory, \ie, employing the same copy-paste strategy as in ABR\cite{liu2023augmented} to paste some old class objects on the new training images.
We follow ABR\cite{liu2023augmented} and set the memory size as 2,000 for the experiments, which stores a selection of instances for all known categories.
As illustrated in \Cref{tab:sup_voc-multi}, with the integration of memory from both original classes and the classes subsequently added from incremental stages, our method shows significant improvement for both old and new classes. Especially in the 10-1 settings, our method achieves a 9.1\% enhancement across all newly introduced classes, increasing from 48.3\% to 57.4\%.
The results effectively highlight that the limitations of our method can be addressed by storing instances in memory.

\begin{table}[ht]
  \centering
  \setlength{\tabcolsep}{4pt}
  \caption{mAP@0.5 results on multiple incremental steps on Pascal-VOC 2007. The best performance in each is presented with \textbf{bold}, and the second best is presented with \underline{underlined}. Methods with $*$ use exemplars.} 
  \vspace{-0.2em}
  \resizebox{0.6\linewidth}{!}{
  \begin{tabular}{l || c c c | c c c}
    \toprule
    \multirow{2}{*}{\textbf{Method}} & \multicolumn{3}{c|}{\textbf{15-1 (6 tasks)}} & \multicolumn{3}{c}{\textbf{10-1 (10 tasks)}}\\
     & \textbf{1-15} & \textbf{16-20} & \cellcolor{Gray}\textbf{1-20}& \textbf{1-10} & \textbf{11-20} & \cellcolor{Gray}\textbf{1-20}\\
    \midrule
    \midrule
    Joint Training &78.0 &70.4 &\cellcolor{Gray}76.1&75.9 &76.3 &\cellcolor{Gray}76.1\\
    Fine-tuning\cite{liu2023augmented} &0.0 &10.5 &\cellcolor{Gray}5.3 &0.0 &5.1 &\cellcolor{Gray}2.6\\
    \midrule
    Faster ILOD\cite{ren2015faster} &66.9 &44.5 &\cellcolor{Gray}61.3 &52.9 &41.5 &\cellcolor{Gray}47.2\\
    MMA\cite{cermelli2022modeling}  &68.3 &54.3 &\cellcolor{Gray}64.1 &59.2 &48.3 &\cellcolor{Gray}53.8\\
    \textbf{BPF (Ours)} & \textbf{71.5}& 53.1 & \cellcolor{Gray}\underline{66.9} &\underline{62.2} &48.3 & \cellcolor{Gray}55.2\\
    \midrule
    ABR*\cite{liu2023augmented} & 68.7 &\underline{56.7} &\cellcolor{Gray}65.7 & 62.0 & \underline{55.7} & \cellcolor{Gray}\underline{58.9}\\
    \textbf{BPF (Ours)*} &\underline{71.3}& \textbf{57.3} & \cellcolor{Gray}\textbf{67.8} &\textbf{62.7} & \textbf{57.4} &\cellcolor{Gray}\textbf{60.1}\\
    \bottomrule
  \end{tabular}}
  \vspace{-0.5em}
  \label{tab:sup_voc-multi}
\end{table}

\section{Additional Analysis for Bridge the Future}
\label{sec:s33}
In Bridge the Future, regions with both high attention scores from feature maps and objectness scores from class-agnostic RPN are excluded from negative samples when training the RoI head, thus maintaining the model's consistency with future stage background definitions. In \Cref{tab:sup_abl_bpf}, we show that both the RPN and spatial attention can independently identify potential foreground objects. And combining them can more accurately identify them, thus performing better.
\begin{table}[h]
  \vspace{-0.9em}
  \centering
  \caption{Ablation Study on selection metrics for potential future class objects in Bridge the Future.}
  \vspace{-0.2em}
  \scalebox{0.9}{
  \begin{tabular}{c c|c c c}
    \toprule
    \textbf{RPN} & \textbf{Attention} & \multicolumn{3}{c}{\textbf{VOC(10-10)}}\\
    \textbf{objectness} & \textbf{scores} & \textbf{1-10} & \textbf{11-20} & \cellcolor{Gray}\textbf{1-20} \\
    \midrule
    \midrule
    &   & 71.2 & 73.3 &\cellcolor{Gray}72.3 \\ 
       &  $\checkmark$  & 71.4 & 73.7 &\cellcolor{Gray}72.6 \\ 
    $\checkmark$ & & 71.3 & 73.7 &\cellcolor{Gray}72.5 \\
    $\checkmark$   & $\checkmark$ & \textbf{71.7} & \textbf{74.0} &\cellcolor{Gray}\textbf{72.9}\\ 
    \bottomrule
  \end{tabular}}
  \vspace{-1.0em}
  \label{tab:sup_abl_bpf}
\end{table}

\section{Detailed Process of Distillation with Future}
\label{sec:s3}

\begin{algorithm}[ht]
	\renewcommand{\algorithmicrequire}{\textbf{Input:}}
	\renewcommand{\algorithmicensure}{\textbf{Output:}}
	\caption{Distillation with Future}
	\label{alg:DwF}
	\begin{algorithmic}[1]
            \REQUIRE Previous model: $\mathcal{M}_{t-1}$; Current model: $\mathcal{M}_{t}$; Current dataset: $\mathcal{D}_t$; Current ground truth $\mathcal{Y}_t$;.
            \STATE Initialization: $\mathcal{L}_{dist}^{roi}=0$
		\STATE Get intermediate model $\mathcal{M}_t^{im}$, trained using $\mathcal{D}_t$ in a fully supervised way
            \FORALL{$\mathcal{I}\in \mathcal{D}_t$}
                \STATE Get proposals $\mathcal{R}$ for distillation, generated by $\mathcal{M}_{t-1}$
                \STATE Divide $\mathcal{R}$ into $\mathcal{R}_1,\mathcal{R}_2\subset \mathcal{R}$, based on their intersection over union with $\mathcal{Y}_t$
                \FORALL{$r_i \in \mathcal{R}_1$}
                    \STATE Get probabilities $p_i^{t-1},p_i^{im},p_i^{t}$ from $\mathcal{M}_{t-1},\mathcal{M}_{t}^{im},\mathcal{M}_{t}$
                    \STATE Reconstruct $M_{t-1}$ background representation: $\hat{p}_i^{c,im} = p_i^{c,im} \times p_i^{b,t-1}$
                    \STATE Get distillation probabilities: $p_i^{dist}=[p_{i}^{\mathcal{C}_{1:t-1},t-1}, \hat{p}_{i}^{\mathcal{C}_{t}\cup\mathcal{B},im}] \in \mathbb{R}^{|\mathcal{C}_{1:t}|+1}$
                    \STATE Get probability distillation loss: $\mathcal{L}_{dist,cls}^{roi}=\mathcal{L}_{KL}(p_i^{dist},p_i^t)$
                    \STATE Get boxes $b_i^{t-1},b_i^{t}$ from $\mathcal{M}_{t-1},\mathcal{M}_{t}$
                    \STATE Get box distillation loss: $\mathcal{L}_{dist,bbox}^{roi}=\mathcal{L}_{2}(b_i^{t-1},b_i^{t})$
                    \STATE Get distillation loss: $\mathcal{L}_{dist}^{roi}=\mathcal{L}_{dist}^{roi}+\mathcal{L}_{dist,cls}^{roi}+\mathcal{L}_{dist,bbox}^{roi}$
                \ENDFOR
                \STATE
                \FORALL{$r_i \in \mathcal{R}_2$}
                    \STATE Get probabilities $p_i^{t-1},p_i^{im},p_i^{t}$ from $\mathcal{M}_{t-1},\mathcal{M}_{t}^{im},\mathcal{M}_{t}$
                    \STATE Reconstruct $M_{t}^{im}$ background representation: $\hat{p}_i^{c,t-1} = p_i^{c,t-1} \times p_i^{b,im}$
                    \STATE Get distillation probabilities: $p_i^{dist}=[\hat{p}_{i}^{\mathcal{C}_{1:t-1},t-1},$ $ p_{i}^{\mathcal{C}_{t},im}, \hat{p}_{i}^{\mathcal{B},t-1}] \in \mathbb{R}^{|\mathcal{C}_{1:t}|+1}$
                    \STATE Get probability distillation loss: $\mathcal{L}_{dist,cls}^{roi}=\mathcal{L}_{KL}(p_i^{dist},p_i^t)$
                    \STATE Get boxes $b_i^{im},b_i^{t}$ from $\mathcal{M}_{t}^{im},\mathcal{M}_{t}$
                    \STATE Get box distillation loss: $\mathcal{L}_{dist,bbox}^{roi}=\mathcal{L}_{2}(b_i^{im},b_i^{t})$
                    \STATE Get distillation loss: $\mathcal{L}_{dist}^{roi}=\mathcal{L}_{dist}^{roi}+\mathcal{L}_{dist,cls}^{roi}+\mathcal{L}_{dist,bbox}^{roi}$
                \ENDFOR
            \ENDFOR
		\ENSURE Distillation with Future loss: $\mathcal{L}_{dist}^{roi}$
	\end{algorithmic}  
\end{algorithm}

\Cref{alg:DwF} shows the detailed process of Distillation with Future strategy. Firstly, we obtain 64 proposals $\mathcal{R}$ for distillation, which are randomly selected out of the top 128 proposals with the highest objectness scores from the RPN network of the old model $\mathcal{M}_{t-1}$. Then we divide the $\mathcal{R}$ into $\mathcal{R}_1,\mathcal{R}_2$ based on the IoU with ground truth labels $\mathcal{Y}_t$. For region $r_i$ in $\mathcal{R}_1$, which is likely to be the region for old classes, we take $\mathcal{M}_{t-1}$ as the primary model and reconstruct its background representation with the model $\mathcal{M}_{t}^{im}$. Then we get the final distillation probabilities: $p_i^{dist}=[p_{i}^{\mathcal{C}_{1:t-1},t-1}, \hat{p}_{i}^{\mathcal{C}_{t}\cup\mathcal{B},im}] \in \mathbb{R}^{|\mathcal{C}_{1:t}|+1}$. On the contrary, for region $r_i$ in $\mathcal{R}_2$, the final distillation probabilities are: $p_i^{dist}=[\hat{p}_{i}^{\mathcal{C}_{1:t-1},t-1},$ $ p_{i}^{\mathcal{C}_{t},im}, \hat{p}_{i}^{\mathcal{B},t-1}] \in \mathbb{R}^{|\mathcal{C}_{1:t}|+1}$.
After getting the probabilities $p_i$ from current model $\mathcal{M}_t$ such that $p_i^t \in \mathbb{R}^{|\mathcal{C}_{1:t}|+1}$, the probability distillation loss is:
\begin{equation}
    \mathcal{L}_{dist,cls}^{roi}=\mathcal{L}_{KL}(p_i^{dist},p_i^t),
\end{equation}
where $\mathcal{L}_{KL}$ represents Kullback-Leibler divergence. Regarding the box distillation, we take the output boxes $b_i^{dist}$ from the old model $\mathcal{M}_{t-1}$ for regions in $\mathcal{R}_1$ and the intermediate model $\mathcal{M}_{t}^{im}$ for regions in $\mathcal{R}_2$. The box distillation loss is:
\begin{equation}
    \mathcal{L}_{dist,bbox}^{roi}=\mathcal{L}_{2}(b_i^{dist},b_i^{t}),
\end{equation}
where $\mathcal{L}_2$ represents the L2 loss.

The final distillation loss in RoI Head is:
\begin{equation}
    \mathcal{L}_{dist}^{roi}=\mathcal{L}_{dist,cls}^{roi}+\mathcal{L}_{dist,bbox}^{roi}.
\end{equation}

\section{Additional Analysis for Distillation with Future}
\label{sec:s4}
Knowledge distillation\cite{hinton2015distilling} serves as a potent strategy in incremental object detection.
Previous methods\cite{liu2023augmented,cermelli2022modeling,joseph2021incremental} transfer the knowledge of the old model to the current model using the current dataset. 
The old model is usually biased toward old classes since it has not trained with new classes, making it hard to extract good representations for new classes. 
However, we find that a number of proposals used in the distillation process do contain the newly introduced classes, and directly distilling the old model's knowledge of these proposals is not wise since the old model has not learned the new classes.

Specifically, \Cref{tab:recall50_DwF} shows that during the second stage of the PASCAL VOC 10-10 setting, the distilled proposals achieve a recall rate of 70.5\% for old classes and \textbf{58.6\% for objects of newly introduced classes}. 
This highlights the importance of our proposed intermediate model, which excels in detecting objects of newly introduced classes.
During the distillation process, we select the old model as the primary distillation model in regions containing old class objects, whereas the intermediate model is chosen as the primary in areas with new class objects. The other model serves as an auxiliary, reconstructing the background probability information of the primary distillation model. This approach provides complementary and comprehensive guidance to the current model.
\begin{table}[ht]
  \centering
  \setlength{\tabcolsep}{4pt}
  \caption{The recall@50 of proposals for distillation generated by the old model's RPN for old classes 1-10 and new classes 11-20. The results are obtained from the second incremental task under the PASCAL VOC 10-10 setting.} 
  \vspace{-0.2em}
  \resizebox{0.7\linewidth}{!}{
  \begin{tabular}{l | c c}
    \toprule
    Classes & old classes (1-10) & new classes (11-20) \\
    \hline
    Recall@50 &70.5  &58.6\\
    \bottomrule
  \end{tabular}}
  \vspace{-0.5em}
  \label{tab:recall50_DwF}
\end{table}